%% file: 0_main.tex
\documentclass[10pt,twocolumn,letterpaper]{article}

\usepackage{iccv}
\usepackage{times}
\usepackage{epsfig}
\usepackage{graphicx}
\usepackage{amsmath}
\usepackage{amssymb}
\usepackage{caption}
\usepackage{subcaption}

\usepackage[pagebackref=true,breaklinks=true,letterpaper=true,colorlinks,bookmarks=false]{hyperref}

\iccvfinalcopy 

\def\iccvPaperID{4339} 

\ificcvfinal\fi

\usepackage{booktabs}
\usepackage{tabularx}
\usepackage{threeparttable}
\usepackage{arydshln}
\usepackage{multirow}
\usepackage[table,xcdraw]{xcolor}
\usepackage[accsupp]{axessibility}  
\usepackage[capitalize]{cleveref}
\crefname{section}{Sec.}{Secs.}
\Crefname{section}{Section}{Sections}
\Crefname{table}{Table}{Tables}
\crefname{table}{Tab.}{Tabs.}

\usepackage[labelsep=period]{caption}

\usepackage{xcolor}
\usepackage{enumitem}

\renewcommand{\paragraph}[1]{\vspace{0.2em}\noindent \textbf{#1 \hspace{0.2em}}}

\definecolor{MyDarkRed}{rgb}{0.66, 0.16, 0.16}
\definecolor{MyDarkBlue}{rgb}{0.16, 0.16, 0.66}
\usepackage{lipsum} 

\begin{document}

\title{Efficient View Synthesis with Neural Radiance Distribution Field}

\author{
Yushuang Wu$^{1,2*}$\quad 
Xiao Li$^{3\dagger}$\quad 
Jinglu Wang$^{3}$\quad
Xiaoguang Han$^{2,1\dagger}$\quad 
Shuguang Cui$^{2,1}$\quad
Yan Lu$^{3}$\\
$^1$FNii, CUHKSZ\quad $^2$SSE, CUHKSZ\quad $^3$Microsoft Research Asia\\
{\tt\small yushuangwu@link.cuhk.edu.cn\quad \{xili11, jinglwa, yanlu\}@microsoft.com\quad}\\{\tt\small \{shuguangcui, hanxiaoguang\}@cuhk.edu.cn}
}

\maketitle
\thispagestyle{empty}

\input{1_abstract}
\footnotetext{$\ast$ This work was done when Yushuang Wu was an intern at MSRA.}
\footnotetext{$\dagger$ Corresponding author.}

\input{2_intro}

\input{3_relatedwork}

\input{4_method}

\input{5_experiments}

\input{6_conclusions}

\clearpage
{\small
\bibliographystyle{ieee_fullname}
\bibliography{ref_new}
}

\end{document}

%% file: 1_abstract.tex
\begin{abstract}
Recent work on Neural Radiance Fields (NeRF) has demonstrated significant advances in high-quality view synthesis. A major limitation of NeRF is its low rendering efficiency due to the need for multiple network forwardings to render a single pixel. Existing methods to improve NeRF either reduce the number of required samples or optimize the implementation to accelerate the network forwarding. Despite these efforts, the problem of multiple sampling persists due to the intrinsic representation of radiance fields.
In contrast, Neural Light Fields (NeLF) reduce the computation cost of NeRF by querying only one single network forwarding per pixel. To achieve a close visual quality to NeRF, existing NeLF methods require significantly larger network capacities which limits their rendering efficiency in practice.
In this work, we propose a new representation called Neural Radiance Distribution Field (NeRDF) that targets efficient view synthesis in real-time. Specifically, we use a small network similar to NeRF while preserving the rendering speed with a single network forwarding per pixel as in NeLF. The key is to model the radiance distribution along each ray with frequency basis and predict frequency weights using the network. Pixel values are then computed via volume rendering on radiance distributions.
Experiments show that our proposed method offers a better trade-off among speed, quality, and network size than existing methods: we achieve a $\sim$254$\times$ speed-up over NeRF with similar network size, with only a marginal performance decline. Our project page is at \href{https://yushuang-wu.github.io/NeRDF/}{yushuang-wu.github.io/NeRDF}.

\end{abstract}


%% file: 2_intro.tex
\section{Introduction}
\label{sec:intro}
\input{figures/fig_pipeline}
The problem of digitally representing a 3D scene for novel view synthesis from arbitrary directions is an important research topic with many applications,
ranging from immersive conferencing to augmented reality.
The breakthroughs made by the pioneering work of NeRF~\cite{mildenhall2020_nerf_eccv20} demonstrate considerable advancements in the view synthesis field, as it represents 3D scenes using implicit radiance fields modeled via neural networks. Despite these advances, a significant drawback of NeRF is its computationally intensive nature, requiring hundreds of network evaluations per pixel, resulting in slow rendering speed (\textit{e.g.} $\sim$0.2 FPS on a high-end GPU). Subsequent research has aimed to enhance the rendering speed by improving importance sampling strategies, reducing the number of samples, or optimizing the code implementation. 
However, these do not address the fundamental problem of multiple sampling inherent in radiance fields, which limits the extent to which rendering speed can be improved (usually 5-30 FPS contingent on implementation) at the cost of increased memory requirements and additional implementation efforts.
To achieve efficient view synthesis, the research community is exploring alternative representations such as neural light fields (NeLF)~\cite{suhail2022light, mildenhall2019_local_tog19, wang2022r2l, sitzmann2021light, attal2021learning}. 
A NeLF maps rays directly into the RGB space, predicting pixel color based on the ray parameters (\textit{e.g.} the origin and direction), thereby reducing the intrinsic computational complexity to one single network forwarding per pixel.
Recent advances have indicated that the synthesis quality of NeLF can be comparable to that of NeRF. However, this usually comes at the cost of much larger networks - for instance, R2L~\cite{wang2022r2l} utilizes an 88-layer MLP network that is 11$\times$ larger than NeRF. The increased size of networks used in NeLF methods results in significantly higher computational and memory costs, as well as limited rendering efficiency in practice.

In this paper, we examine the challenge of achieving efficient view synthesis in practical settings, which requires a careful balance among multiple considerations including perceptual quality (measured by PSNR), computational efficiency (measured by FPS), and memory requirements (\textit{e.g.} model size).
To achieve this goal, we start from a key observation: previous NeLF methods have no explicit perception to the 3D geometry information. As known, NeRF attends to the radiance information at spatial locations along each camera ray as illustrated in \cref{fig:pipeline}, which enables NeRF to have a perception of the 3D geometry information of a scene. However, NeLF learns the direct mapping from the huge ray space to the pixel RGB space without any 3D prior. 
This paradigm hinders NeLF from learning the actual intrinsic 3D layout/structure of a scene, which results in an over-dependence on a large network capacity, as previous NeLF methods~\cite{attal2021learning, wang2022r2l} suffer from.

Based on this observation, we propose a novel implicit representation, termed as \textbf{Neural Radiance Distribution Field (NeRDF)}. NeRDF is based on a simple yet effective key idea: from the input ray space as in NeLF, NeRDF yet learns the radiance distribution of a given ray, so that the spatial 3D geometry information can be perceived.
Specifically, we train the network to produce the radiance distribution along a ray, parameterized using a set of trigonometric functions. 
The final pixel color is re-synthesized via volume rendering from the output radiance distribution as in NeRF. 
Thus, the proposed NeRDF combines the strengths of both NeRF-based and NeLF-based methods. NeRDF models the parametric ray radiance distribution in order for a significantly more compact target space than direct modeling the ray-to-pixel mapping. This enables NeRDF to represent scenes with much smaller neural networks that are comparable to those in NeRF. 
Additionally, the prediction of the radiance distribution of a ray only takes one single network forwarding, so only one evaluation is required to render a pixel as in NeLF.

The learning of NeRDF is based on a knowledge distillation framework inspired by~\cite{wang2022r2l}, where a teacher NeRF synthesizes dense and diverse views that then serve as the pseudo-training data. 
We further contribute three novel designs to enhance the framework: (i) an input ray encoding method that captures rich ray information, (ii) an online view sampling strategy that expands the diversity of the pseudo-training data, and (iii) a volume density constraint loss that promotes the learning of a strong 3D prior.

We have evaluated the efficiency of the NeRDF method on the Real Forward-Facing (LLFF) dataset~\cite{mildenhall2020_nerf_eccv20}. On average, without any specific optimization in network inference, our proposed NeRDF method achieves a comparable visual quality to existing methods while rendering at a speed of $\sim$21 FPS and using only an 8-layer MLP network. This is $\sim$5$\times$ faster than the R2L method~\cite{wang2022r2l} and $\sim$100$\times$ faster than the original NeRF. Additionally, our method can benefit from any off-the-shelf inference optimization methods. By employing tiny-cuda-nn~\cite{muller2021real} as our inference backend, our NeRDF achieves a rendering speed of $\sim$369 FPS, which is a $\sim$1400$\times$ speed-up over an unoptimized NeRF and a 10-15$\times$ speed-up compared with previous methods using the same backend.
We also demonstrate that the NeRDF method provides a superior trade-off between speed, memory, and visual quality compared with previous NeRF-based and NeLF-based methods.
Our contributions are as follows:
\begin{itemize}[noitemsep]
\vspace{-1mm}
    \item A new neural representation for 3D scenes, Neural Radiance Distribution Field (NeRDF), that outputs radiance distribution along rays.  
    \item A method for NeRDF learning with a compact neural network that has high rendering speed, low memory cost, and plausible quality.
    \item An efficient view synthesis solution that has a good trade-off among visual quality, speed, and memory.
\end{itemize}
\input{tables/representation_compare}


%% file: figures/fig_pipeline.tex
\begin{figure*}[tb]
    \centering
    \includegraphics[width=0.95\textwidth]{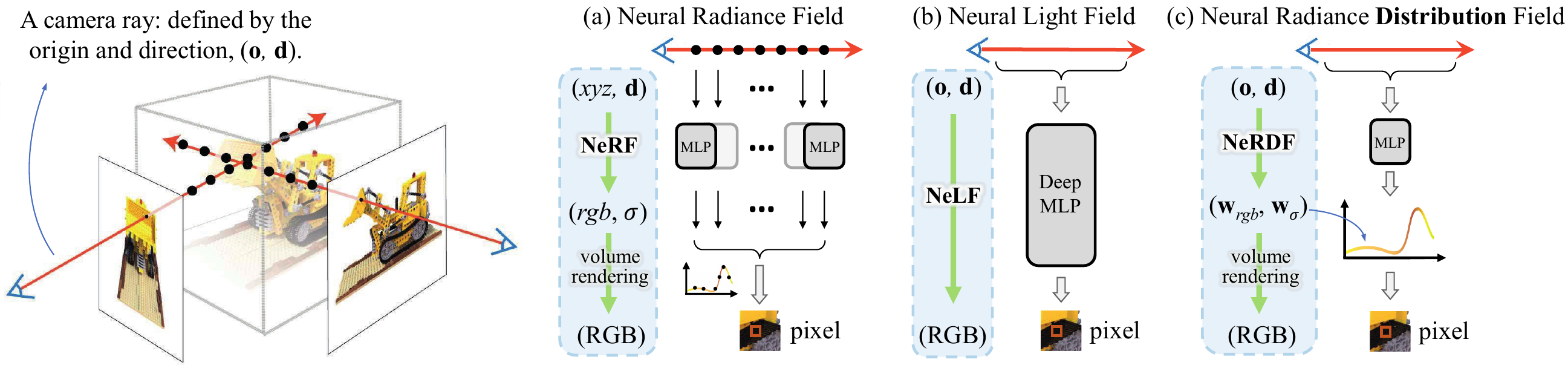}
    \vspace{-0.5mm}
    \caption{The overview of our Neural Radiance Distribution Field (NeRDF) and the comparison of (a) NeRF, (b) NeLF, and (c) NeRDF. NeRF requires hundreds of network forwarding per ray to predict the volume density and color, and output the pixel RGB via volume rendering. NeLF takes only one single forwarding per ray to predict the pixel RGB but strongly depends on a much larger network. Our NeRDF absorbs both advantages that takes only one single forwarding per ray as NeLF with only a small network as NeRF. The key idea is to directly predict the radiance distribution from the ray input. } 
    \vspace{-5mm}
    \label{fig:pipeline}
\end{figure*}

%% file: tables/representation_compare.tex
\begin{table*}[tb]
\centering
\caption{Comparison between different neural representations. As show, previous methods struggle at fulfilling low memory, high speed, and high quality at the same time. Our method breaks this impossible trinity.}
\label{tab:representation_compare}
\resizebox{0.90\linewidth}{!}{
\begin{tabular}{@{}cccccc@{}}
\toprule
               & \multicolumn{2}{c}{NeRF-based~\cite{yu2021_plenoctrees_arxiv, fang2021neusample, neff2021_donerf_egsr21,  kurz2022adanerf,muller2021real, muller2022instant}}      & \multicolumn{2}{c}{NeLF-based~\cite{wang2022r2l, attal2021learning, wang2022progressively}}                                 &      \\ \cmidrule(lr){2-3} \cmidrule(lr){4-5}
\multirow{-2}{*}{Methods} & Hybrid NeRF                 & Accelerated Backend           & Small Network & Large Network               & \multirow{-2}{*}{\textbf{NeRDF (Ours)}} \\ \midrule
Visual Quality & High                         & High & {\color[HTML]{CB0000} Low} & High                        & \textbf{High} \\
Rendering Speed           & {\color[HTML]{CB0000} Medium} & {\color[HTML]{CB0000} Medium} & Fast          & {\color[HTML]{CB0000} Slow} & \textbf{Very Fast}                      \\
Memory Cost    & {\color[HTML]{CB0000} Large} & Low  & Low                        & {\color[HTML]{CB0000} High} & \textbf{Low}  \\ \bottomrule
\end{tabular}
}
\vspace{-4mm}
\end{table*}

%% file: 3_relatedwork.tex
\section{Related Work}
\label{sec:related_works}
\paragraph{Neural 3D representations.}
Recently, the neural field representation of 3D scenes has attracted significant attention from the literature~\cite{mildenhall2020_nerf_eccv20, sitzmann2019scene, mescheder2019occupancy, wang2021neus}. These techniques utilize multi-layer perceptrons to generate implicit fields such as sign distance functions or volume radiance fields.
Of particular note is the Neural Radiance Fields (NeRF)~\cite{mildenhall2020_nerf_eccv20, barron2021mip, barron2022mip}. NeRF has demonstrated its effectiveness on the task of view synthesis from a limited number of input views, leading to an explosion number of follow-up works that extend its capabilities to other tasks. These include the handling of dynamic scenes~\cite{du2021_nerflow_iccv21, xian2021_space_cvpr21, park2021_nerfies_iccv21, pumarola2021_dnerf_cvpr21, park2021hypernerf}, human digitization~\cite{te2022neural, hong2022headnerf, su2021nerf, peng2021animatable, liu2021neural}, shape and appearance modeling~\cite{deng2021deformed, li2022estimating, zhang2021nerfactor, bi2020neural}, 3D-aware synthesis~\cite{chan2021pi, deng2022gram}, and many others. A comprehensive overview of NeRF and its related applications can be found in~\cite{xie2022neural}.
Our method aims to train a succinct neural representation, which will allow for more efficient view synthesis and thereby benefit many of NeRF's applications.

\paragraph{Efficient view synthesis.}
To improve the rendering efficiency of NeRF, numerous existing works have reduced the number of queried samples along rays through the use of auxiliary data structures, such as octrees~\cite{liu2020_nsvf_nips20, yu2021_plenoctrees_arxiv},
hash-tables~\cite{muller2022instant}, the application of additional networks for importance sampling~\cite{lindell2021_autoint_cvpr21, neff2021_donerf_egsr21, fang2021neusample, kurz2022adanerf}, or pre-baking the radiance field~\cite{garbin2021_fastnerf_arxiv}.
These methods accelerate NeRF to some degree, but come with additional computational and memory costs. Concurrently, tiny-cuda-nn~\cite{muller2021real} attempts to increase the inference speed of shallow MLPs through dedicated optimization at the CUDA level. However, they do not address the fundamental issue of multiple sampling.
An alternative to these approaches is the use of other neural representations, with the neural light field (NeLF) becoming a popular choice as it requires only a single evaluation per pixel~\cite{wang2022r2l, attal2021learning, wang2022progressively}.
While NeLF-based methods can produce high-quality view synthesis results comparable to NeRF, they also entail significant memory costs due to their deeper networks used~\cite{wang2022r2l} or additional networks to map rays into latent spaces~\cite{attal2021learning}. Despite these high memory costs, their rendering efficiency still remains limited with larger networks.
Our method focuses on efficient view synthesis with a straightforward design, carefully balancing visual quality, speed, and memory costs.

\paragraph{Knowledge distillation.}
Knowledge distillation (KD), as commonly used in the field of network compression, involves training a small model (known as the student) to mimic the outputs and intermediate feature representations of a larger pre-trained model (referred to as the teacher)~\cite{hinton2015distilling, chen2017learning, passalis2018learning, zagoruyko2016paying, tian2019contrastive}.
In the context of view synthesis, the R2L method~\cite{wang2022r2l} exploits KD by transferring knowledge from a pre-trained teacher NeRF network to a student NeLF network, by matching the rendered pixel value between the two models~\cite{wang2022r2l}.
Our proposed approach extends the knowledge distillation process outlined in~\cite{wang2022r2l} by incorporating an additional matching between the volume density distribution of the target NeRDF and the teacher NeRF.

%% file: 4_method.tex
\section{Method}%
\label{sec:method}
In this section, we first provide the preliminary of view synthesis based on neural implicit representations including NeRF and NeLF in \cref{sec:sub:preliminary}. Next, we introduce our proposed representation of NeRDF in \cref{sec:sub:nerdf}. Lastly, we detail the learning process of a compact NeRDF via distillation from a teacher NeRF in \cref{sec:sub:train_nerdf}.

\subsection{Preliminary}%
\label{sec:sub:preliminary}
\paragraph{View synthesis with neural fields.}
Given sparse multi-view images as observations of a 3D scene, the objective is to train a neural network that implicitly captures the scene's structure, which allows for the synthesis when given an arbitrary, unseen view of the scene.

\paragraph{Neural Radiance Fields (NeRF).}
We begin by reviewing the fundamentals of Neural Radiance Fields (NeRF) as described in~\cite{mildenhall2020_nerf_eccv20}. 
NeRF is a Multi-layer Perceptron (MLP) network that maps 3D coordinates of a point to its radiance and volume density values. Mathematically, this can be represented as the following function:
\begin{equation}
    F_{\text{NeRF}}: (x, y, z, \theta, \phi) \rightarrow (\mathbf{c}, \sigma),
    \label{equ:nerf_mapping}
\end{equation}
where $(x,y,z)$ are the spatial coordinates, $(\theta, \phi)$ represent the 2D viewing direction,  $\mathbf{c} = (r, g, b)$ is the emitted color, and $\sigma$ is the volume density.
Given a camera ray defined by $\mathbf{r}(t)=\mathbf{o}+t\mathbf{d} \in \mathbb{R}^{3}$, where $\mathbf{o}$ and $\mathbf{d}$ are the origin and direction of the ray respectively, the rendered color $C(\mathbf{r})$ can be calculated with volume rendering:
\begin{equation}
    C(\mathbf{r}) = \int_0^{\infty} T(t)~\sigma(\mathbf{r}(t))~\mathbf{c}(\mathbf{r}(t), \mathbf{d})dt,
    \label{equ:nerf_render}
\end{equation}
where
\begin{equation}
T(t) = \exp\left(-\int_0^t \sigma(\mathbf{r}(t))ds \right)
\label{equ:transmit}
\end{equation} 
In practice, the integration of~\cref{equ:nerf_render} is approximated discretely through Monte Carlo sampling of points along the camera ray. The network $F_{\text{NeRF}}$ is queried at each of these points, providing predictions of $\sigma(\mathbf{r}(t))$ and $\mathbf{c}(\mathbf{r}(t), \mathbf{d})$ using~\cref{equ:nerf_mapping}. 
It is typical that a NeRF requires approximately 100-200 network queries to compute the RGB pixel color value $C(\mathbf{r})$ for a given camera ray $\mathbf{r}$.

\paragraph{Neural Light Fields (NeLF).}
To reduce computational complexity, a scene can be also represented as a Neural Light Field (NeLF). Different from NeRF, a NeLF directly maps a 4D-oriented camera ray to the RGB color space, as described in \cite{attal2022learning}:
\begin{equation}
    F_{\text{NeLF}}: \mathbf{r} \in \mathbb{R}^4 \rightarrow \mathbf{c} \in \mathbb{R}^3.
    \label{equ:nelf_mapping}
\end{equation}
Thus, NeLF simplifies the rendering process by eliminating the need for analyzing 3D radiance information and conducting volume rendering, requiring only a single network evaluation to compute the pixel color of a given ray.
However, the learning of a NeLF that produces high-quality synthesis greatly challenges the network capacity. For example, the state-of-the-art NeLF-based method R2L, described in \cite{wang2022r2l}, uses an 88-layer MLP network (11$\times$ deeper than in NeRF), resulting in a rendering speed of only $\sim$5 FPS.
Reducing the depth of the neural network for R2L leads to severe quality degradation, as evidenced in the results of \cite{wang2022r2l} and our observations as shown in \cref{fig:qualitative_compare}.

\subsection{Neural Radiance Distribution Field}
\label{sec:sub:nerdf}
\cref{tab:representation_compare} summarizes the challenge in achieving a balance among the quality, rendering efficiency, and memory cost for existing neural field methods. Our goal is to overcome this challenge and enable high-quality, real-time view synthesis with low memory overhead (network size).

\paragraph{3D prior.}
In the comparison between NeRF and NeLF, a key factor contributing to NeRF's ability to achieve high-quality view synthesis with a smaller network than NeLF is its reliance on a 3D prior. 
The multi-view consistency and 3D geometric information of a scene are naturally taken into account during perceiving the radiance information at each spatial location, resulting in a compact representation with small networks.
However, this advantage comes at the cost of requiring numerous evaluations to compute the color of each pixel via volume rendering in \cref{equ:nerf_render}. 
In contrast, NeLF directly maps from ray space to RGB space, without considering any spatial radiance information. This leads to a strongly view-dependent representation, with each pixel from different views being addressed independently without any 3D prior information. 
Therefore, larger networks are typically required to memorize all views for synthesis, as seen in~\cite{wang2022r2l}.

\paragraph{Neural Radiance Distribution Field (NeRDF).}
Different from NeRF and NeLF, our proposed representation, the Neural Radiance Distribution Field (NeRDF), takes the camera ray of each pixel as input and maps it into the radiance distribution along that ray:
\begin{equation}
    F_{\text{NeRDF}}: \mathbf{r} \rightarrow (\mathbf{w}^C, \mathbf{w}^\sigma),
    \label{equ:nerdf_mapping}
\end{equation}
where $\mathbf{w}^C$ and $\mathbf{w}^\sigma$ are a set of parameters that define the radiance opacity and color distributions for the given ray $\mathbf{r}$. The pixel color is then computed via volume rendering (as in \cref{equ:nerf_render}) by a dense sampling from the parametric radiance distributions.

\cref{fig:pipeline} shows the overall pipeline of NeRF, NeLF, and our proposed NeRDF. 
Our proposed NeRDF had the following advantages:
(i) As in NeRF, NeRDF is built upon the volume rendering which obtains the pixel color through the ray marching of radiance values. This gives the neural representation the ability to model the 3D prior, which significantly narrows down the vast target space of NeLF, allowing for a smaller network capacity required during the learning process.
(ii) As in NeLF, the input of NeRDF is defined in the ray space. As a result, only a single network evaluation is required to output the full radiance distribution of a ray, reducing the computation cost of multiple queries.

\paragraph{Parameterize radiance distribution.}
The radiance opacity/color distribution along a ray is generally arbitrary. To parameterize it, we conduct discrete Fourier analysis by integrating pre-defined trigonometric functions with discrete frequencies. 
The opacity and color function of a ray, $\sigma(t)$ and $C(t)$, respectively, can be represented as follows:
\begin{align}
\label{equ:dft_approx1}
\sigma(t) &= \sum_{i=0}^{2K-1} w_i^{\sigma} \cdot \mathcal{T}_i(t), \\
\label{equ:dft_approx2}
C(t) &= \sum_{i=0}^{2K-1} w_i^{C} \cdot \mathcal{T}_i(t), 
\end{align}
where $t$ represents the distance stamp along the ray, $w_i^{\sigma(C)}$ refers to the coefficients predicted by NeRDF as outlined in \cref{equ:nerdf_mapping}, $K$ is the number of frequencies. The frequency basis $\mathcal{T}_i(t)$ is defined as:
\begin{align}
\mathcal{T}_i(t) = 
\left\{
    \begin{aligned}
    &\mathrm{cos}(\frac{i\pi}{T}t) & \text{for even $i$,} \\
    &\mathrm{sin}(\frac{(i+1)\pi}{T}t) & \text{for odd $i$.}
    \end{aligned}
\right.
\label{equ:dft_tri}
\end{align}
We use an MLP network to predict $\mathbf{w}^{\sigma}$ and $\mathbf{w}^{C}$ for the volume density distribution and the radiance color distribution, respectively. In total, this leads to an output with $8K$ channels. 
As demonstrated in \cref{fig:mix_gaussian}, we show an example of the volume density distribution of a ray produced by NeRF (left) and NeRDF (right). The results show that NeRDF reproduces a similar radiance distribution to NeRF's.
\input{figures/fig_mix_gaussian}

\subsection{NeRDF Learning}
\label{sec:sub:train_nerdf}
Given a set of multi-view posed images as training data, the proposed NeRDF is trained by minimizing an $L_2$ norm loss function between the synthesized RGB values, $\hat{C}(\mathbf{r})$, and the ground-truth RGB values, $C(\mathbf{r})$. The loss function is defined as follows:
\begin{equation}
    L_{render} = \sum_{\mathbf{r}} ||\hat{C}(\mathbf{r}) - C(\mathbf{r})||_2^2
    \label{equ:render_loss}
\end{equation}
The predicted RGB value, $\hat{C}(\mathbf{r})$, is calculated by performing two steps: (i) sampling multiple points along the ray and obtaining their opacity and color from the parametric radiance distribution using \cref{equ:dft_approx1} and \cref{equ:dft_approx2}, and (ii) computing the volume integration with \cref{equ:nerf_render}.

To learn a high-quality NeRDF representation with compact neural networks, there are still several technical challenges:
(i) small changes in the input ray space can result in significant variations in the output RGB values, particularly at the edges of foreground objects.
(ii) the training views are not sufficient to learn a NeRDF with an input defined in the ray space.
(iii) there is a need to constrain the predicted radiance distributions of correlated rays to guarantee a valid 3D prior is learned. We address these challenges with the following novel designs.

\paragraph{Input ray encoding.} To tackle the first challenge, we adopt a strategy of compound encoding to project the input into a higher-dimensional space, thereby incorporating more complex information about the rays.
Given a ray $\mathbf{r}(t) = \mathbf{o} + t\mathbf{d}$, we take into account the different impacts of the origin and direction on the viewed pixels, and encode $\mathbf{o}$ and $\mathbf{d}$ differently. 
Specifically, we utilize the typical frequency-based positional encoding method as in NeRF \cite{mildenhall2020_nerf_eccv20} for encoding the origin, and a spherical-harmonic (SH) based encoding inspired by~\cite{muller2021real, wang2022fourier} for the direction, where the coefficients of the SH functions are used as the embedding vector of the direction. 
To include the path information of the input ray, we additionally sample $N$ points along the ray $\mathbf{r}$. This is achieved by conducting a stratified sampling technique as described in \cite{wang2022r2l}. Finally, we embed this $3N$-dimensional vector using the same frequency-based positional encoding technique as applied in \cite{mildenhall2020_nerf_eccv20} and concatenate it with the encoded origin and direction vector, forming the final input.

\paragraph{Online view sampling.} To tackle the second challenge of only sparse views available for NeDRF learning, we utilize a teacher NeRF to synthesize pseudo data in the form of numerous views. 
In the recent work of R2L \cite{wang2022r2l}, a teacher NeRF is employed to pre-produce around 10,000 pseudo images with sufficient view coverage for offline training. However, we adopt a more efficient approach called online view sampling (OVS), which involves randomly sampling camera poses for pseudo-data generation during training. Besides the improved training efficiency, OVS also provides two additional benefits: (1) a wider range of sampled views can be obtained, and (2) not only the final pixel RGB but also the opacity and color information along the ray can be leveraged to improve the learning of 3D prior.

\paragraph{Volume density constraint.} To alleviate the issue of inconsistent geometry across different views when individually fitting the radiance distribution along each ray, a volume density constraint (VDC) loss is introduced. 
The VDC loss minimizes the $L_2$ norm distance of the volume density at each point between two models, the student NeRDF and teacher NeRF:
\begin{equation}
    L_{vdc} = \lambda\sum_{\mathbf{r}} \sum_{t} ||\hat{\sigma}_1(t) - \hat{\sigma}_2(\mathbf{r}(t))||_2^2,
    \label{equ:density_loss}
\end{equation}
where $\lambda$ is the weight for balancing the VDC loss, $\hat{\sigma}_1(t)$ and $\hat{\sigma}_2(\mathbf{r}(t))$ denote the normalized volume density at the same point predicted by NeRDF (\cref{equ:dft_approx1}) and NeRF, respectively.
The proposed VDC not only ensures multi-view consistency in NeRDF, but also extends the knowledge distillation from using the final pixel RGB only to additionally involving the ``intermediate" features, the volume density. 
This is in line with common practices in other knowledge distillation tasks for improving performance, as noted in \cite{chen2017learning}.

\paragraph{Implementation details.}
The proposed method is implemented using the PyTorch framework~\cite{paszke2017pytorch}. An 8-layer MLP with a width of 384, unless specified otherwise, is used to build a NeRDF. 
The MLP network structure is similar to that of R2L~\cite{wang2022r2l}, with skip connections added between neighboring layers.
The parameter $K$ in the Fourier trigonometric function approximation is set to 12.
The volume rendering in \cref{equ:nerf_render} is approximated using 64 uniform samples from the radiance distribution, and the sampling process was implemented with Taichi~\cite{hu2019taichi}. 
The input ray encoding uses $N=16$ and 8 degrees in the SH embedding to encode the ray direction $\mathbf{d}$ into a 64-dimensional vector.
The positional encoding uses 10 frequencies, as in NeRF. 
The OVS training uses random sampling of 1 view origin and 2,048 directions to form a batch of training rays. The weight $\lambda$ in the VDC loss of \cref{equ:density_loss} is set to 0.1. 
The entire NeDRF training was conducted on a single RTX3090 GPU with a learning rate of $5\times10^{-4}$ for 400k-600k iterations. Other implementation details are provided in supplementary materials. 


%% file: figures/fig_mix_gaussian.tex
\begin{figure}[ht] \centering
    \includegraphics[width=0.49\textwidth]{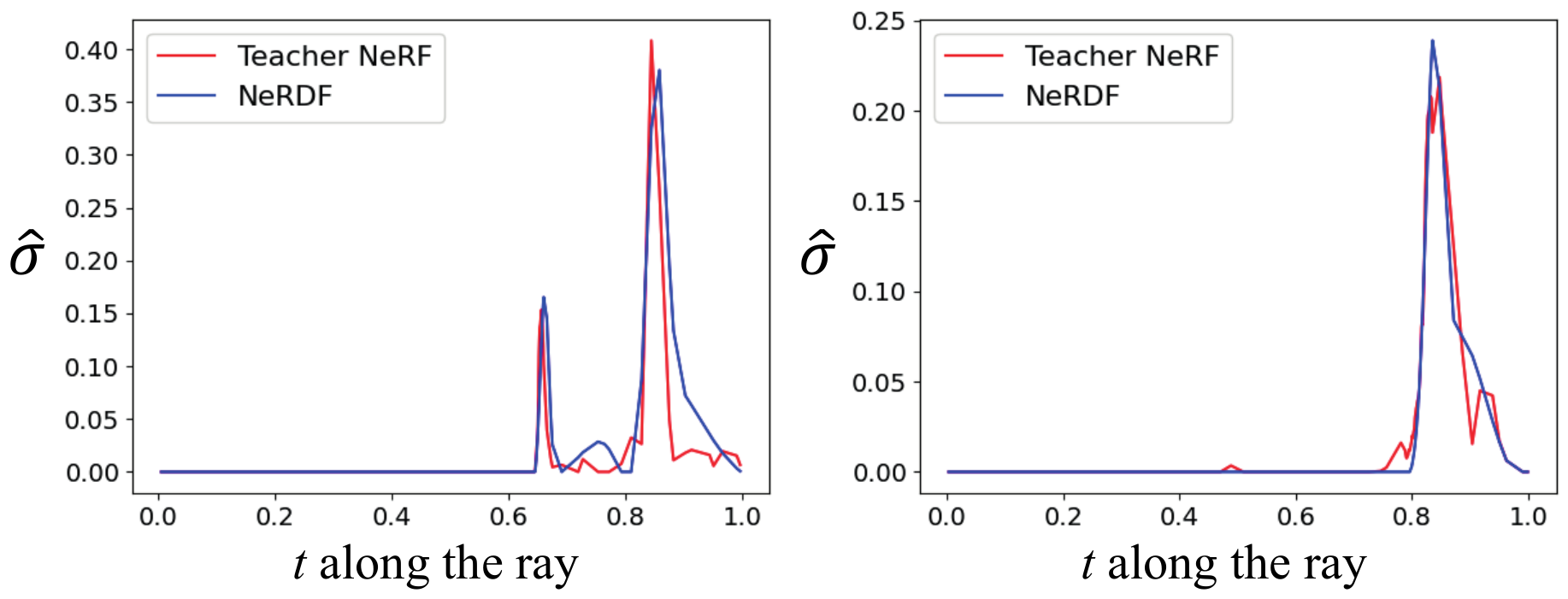}
    \\
    \caption{
        The radiance distribution (normalized opacity) predicted by NeRDF (blue) compared with NeRF (red).
        Overall, the output of NeRDF faithfully reproduces the distribution shape from NeRF.
    } 
    \vspace{-3mm}
    \label{fig:mix_gaussian}
\end{figure}

%% file: 5_experiments.tex
\section{Experiments}%
\label{sec:Experiments}
To demonstrate the advantages of our proposed NeRDF, we perform comparisons against multiple NeRF-based and NeLF-based efficient view synthesis methods in \cref{sec:sub:comparison}.
We show that the proposed NeRDF leads to a better trade-off among speed, memory cost, and quality.
We also perform a set of ablation studies in \cref{sec:sub:ablation_study} to validate our proposed NeRDF as well as our training design.

\subsection{Experiment Setup}
\label{sec:sub:expr_setup}
\paragraph{Datasets.}
We conduct experiments on the Real Forward-Facing (LLFF) dataset~\cite{mildenhall2020_nerf_eccv20}.
The LLFF dataset consists of 8 scenes captured with a handheld cellphone.
The number of images for each data ranges from 20 to 62.
Following \cite{wang2022r2l} and \cite{lindell2021_autoint_cvpr21}, we downsample all images into the resolution of 504$\times$378 pixels in both training and test. Results on a higher resolution (1008$\times$756) are provided in the supplementary materials.

\paragraph{Metrics.}
To better analyze the trade-off between different factors, we measure 
(1) the PSNR (dB) for evaluating the synthesis quality;
(2) the network size (and the memory cost of additional data structure, if any) for evaluating the memory cost;
(3) the running time in FPS for evaluating the rendering speed.

\paragraph{Baselines.}
We compare our method with the following efficient view synthesis methods based on NeRF: TermiNeRF~\cite{piala2021terminerf}, DONeRF~\cite{neff2021_donerf_egsr21}, AdaNeRF~\cite{kurz2022adanerf}, NeX~\cite{wizadwongsa2021nex}, and Instant-NGP~\cite{muller2022instant}. We also compare our method with NeLF-based methods including
RSEN~\cite{attal2021learning} and R2L~\cite{wang2022r2l}. 
We do not compare our method with baking-based methods such as \cite{yu2021plenoxels, garbin2021_fastnerf_arxiv, Reiser2021_kiloNeRF_iccv21}, as they target speed-up NeRF by incurring a high memory cost. Our method, on the other hand, aims to keep the memory cost as low as possible.

\subsection{Main Results}
\input{figures/fig_qualitative_compare}
\label{sec:sub:comparison}
\paragraph{Comparison with NeLF-based methods.}
The quantitative results of our method and NeLF-based methods are listed in \cref{tab:res_llff_nelf}. These results are averaged over 8 scenes of the LLFF dataset and include synthesis quality, speed, and memory cost, evaluated on an RTX3090 GPU.
Our method outperforms RSEN~\cite{attal2021learning} in terms of rendering speed,
achieving a speed-up of roughly 6-10 times while maintaining similar or superior visual quality.
In terms of synthesis quality, both our method and R2L~\cite{wang2022r2l} produce high-quality results when using large networks (R2L-88 v.s. NeRDF-48). However, R2L struggles to produce results of adequate quality with smaller networks (R2L-16), whereas our method can still generate plausible results even with an 8-layer network (NeRDF-8). This makes our method possible to perform real-time, high-quality view synthesis with low memory cost. The advantage of our method in rendering speed with a small network is further demonstrated in the PSNR-FPS trade-off curve in \cref{fig:psnr_curve_nelf}, where our method outperforms NeLF-based methods in the high-FPS region.
\input{tables/results_llff_nelf}
\input{figures/fig_psnr_curve_nelf}

\paragraph{Comparison with NeRF-based methods.}
The quantitative results of our method and NeRF-based methods on the LLFF dataset are presented in \cref{tab:res_llff_nerf}. Our NeRDF-8 has already exhibited significant speed advantages and outperformed almost all NeRF-based methods, except for NeX~\cite{wizadwongsa2021nex}. However, NeX has a higher memory cost due to its storage of multi-plane feature maps and a higher computational cost (in MFLOPs) of 42.71 compared with our 2.60. The PSNR-FPS trade-off curves of NeRF-based methods and our NeRDF are illustrated in \cref{fig:psnr_curve_nerf}, where our method demonstrates a significantly better trade-off than previous NeRF-based methods. 
\input{tables/results_llff_nerf}
\input{figures/fig_psnr_curve_nerf}

\paragraph{Qualitative results.}
Finally, we conduct a qualitative comparison of our view synthesis results with other methods in \cref{fig:qualitative_compare}.
The results indicate that our method (NeRDF-8, \cref{fig:qualitative_compare}e) produces plausible visual quality, although it has a slightly lower PSNR compared with other NeRF-based methods. 
The comparison also shows that previous NeLF-based methods, such as R2L in \cref{fig:qualitative_compare}d, fail to synthesize acceptable visual quality under similar speed requirements and memory cost constraints.

\subsection{Discussions}
\label{sec:sub:ablation_study}
In this section, we perform a set of ablation studies on the FERN data, unless specified otherwise, from the LLFF dataset~\cite{mildenhall2020_nerf_eccv20}
to validate the key components of NeRDF and our training strategy.

\paragraph{Inference time breakdown.}
Although NeRDF only requires a single network forward pass per pixel, it includes the ray-marching process for pixel color rendering. \cref{tab:time_analysis} displays the breakdown of inference time for NeRDF-8. The results indicate that the majority of the inference time is still devoted to network inference. The additional time required for volumetric rendering only accounts for 5.3\% of the total runtime for an 8-layer MLP. This highlights our advantage over NeRF-based methods, as we eliminate the most time-consuming part of multiple network forward passes while preserving the volumetric rendering component for synthesis quality. Besides, the training time and the optimization details are provided in supplementary materials. 
\input{tables/time_analysis}

\input{tables/ablation_OVS_VDC}
\paragraph{NeRDF components.}
The proposed NeRDF method outperforms the NeLF-based method in representing scenes with small network capacities by incorporating three designs: outputting radiance distribution, incorporating online view sampling, and enforcing a view density constraint.
To validate these designs, we conducted ablation experiments on two data, FERN and FLOWER, from the LLFF dataset. We start from an 8-layer MLP in R2L~\cite{wang2022r2l} (netwidth is 256) and gradually add our components to observe the impact of each component.
The results, shown in \cref{tab:ablation_ovs_vdc}, demonstrate that using radiance distribution as the output (\cref{tab:ablation_ovs_vdc}b) leads to 1.4-2.0 dB improvement compared with directly predicting pixel RGB values (\cref{tab:ablation_ovs_vdc}a).
The online view sampling (OVS) further enhances the performance of NeRDF (\cref{tab:ablation_ovs_vdc}c) with 0.9-2.8 dB improvement.
Finally, introducing the view density constraint (VDC) increases the performance by approximately 0.1 dB (\cref{tab:ablation_ovs_vdc}d).
As seen in \cref{fig:sigmaloss_effect_disparity}, the disparity maps generated by NeRDF with VDC are more reasonable compared with those generated without VDC, further validating the benefits of the proposed VDC for view-consistency. 
These results indicate that NeRDF effectively learns the 3D prior and provides a better scene representation than NeLF with a compact neural network.
\input{figures/fig_volume_density_disparity}

\paragraph{The number of Fourier frequencies.}
The radiance distributions in NeRDF are approximated using a group of trigonometric functions (TrigF) through discrete Fourier transformations. 
We analyze the impact of the number of frequencies in TrigF (denoted as $K$), with results listed in~\cref{tab:ablation_K_GMM}. We find the best $K$ is 12, and the performance of NeRDF is robust to the choice of $K$ ranging from 4 to 24. We also provide the results of using a Gaussian Mixture Model as the alternative in supplementary materials.
\input{tables/ablation_K_GMM}
\input{tables/ablation_Input_Encode}

\paragraph{Input ray encoding.}
We analyze various options for encoding input rays in \cref{tab:ablation_input_encode}. As shown, it results in a decrease of 1.1dB without encoding points sampled along the ray's path (as used in~\cite{wang2022r2l}), as shown in \cref{tab:ablation_input_encode}a. Besides, encoding both the direction and on-the-path points (\cref{tab:ablation_input_encode}c) results in a 0.1dB improvement compared with encoding points only (\cref{tab:ablation_input_encode}b). Omitting the direction encoding, as depicted in \cref{tab:ablation_input_encode}d, results in a drop of $\sim$0.1dB compared with using all three components, as shown in \cref{tab:ablation_input_encode}e.

\paragraph{Limitations and Future works.}
Our method is not without limitations. 
NeRDF requires further expansion in order to effectively handle 360-degree scenes with high fidelity. As future avenues of research, we propose to further improve the view-synthesis quality as well as extend NeRDF to handle dynamic scenes.

%% file: figures/fig_qualitative_compare.tex
\begin{figure*}[ht!] \centering
    \makebox[0.192\linewidth]{\scriptsize (a) Teacher NeRF (27.75 dB)}
    \makebox[0.192\linewidth]{\scriptsize (b) NeX (27.26 dB)}
    \makebox[0.192\linewidth]{\scriptsize (c) RSEN (27.94 dB)}
    \makebox[0.192\linewidth]{\scriptsize (d) R2L-16 (24.42 dB)}
    \makebox[0.192\linewidth]{\scriptsize (e) \textbf{NeRDF-8} (26.53 dB)}
    \\
    \includegraphics[width=0.192\linewidth]{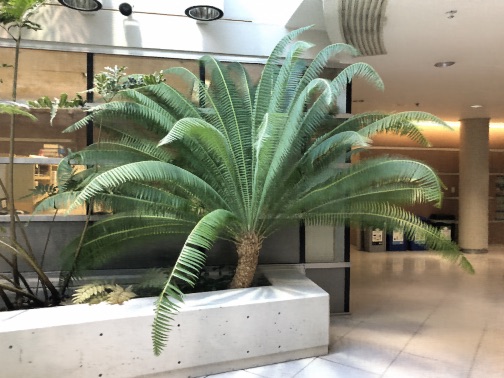}
    \includegraphics[width=0.192\linewidth]{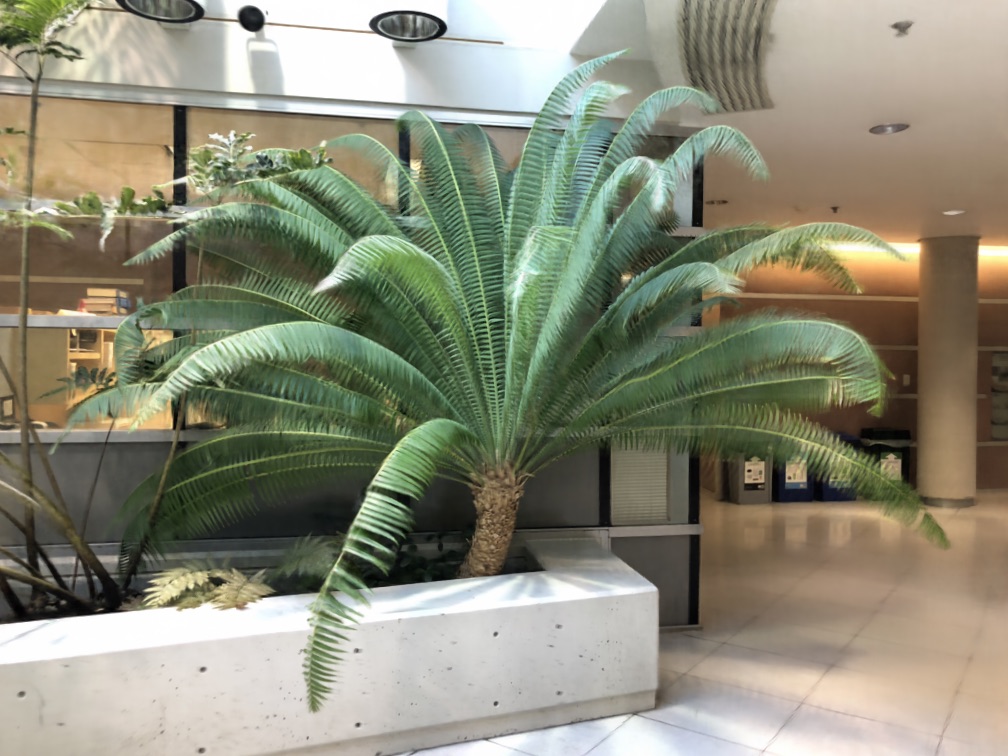}
    \includegraphics[width=0.192\linewidth]{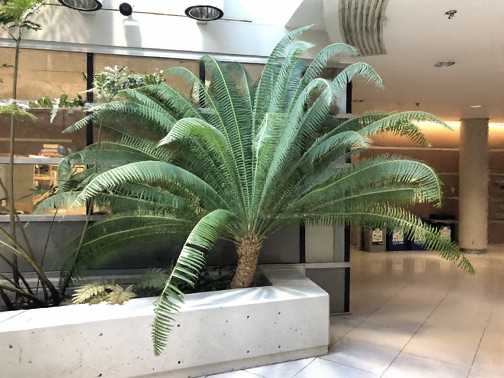}
    \includegraphics[width=0.192\linewidth]{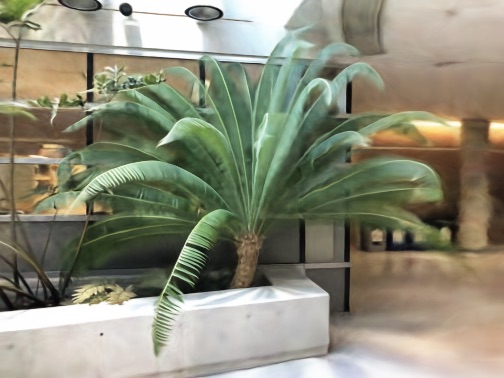}
    \includegraphics[width=0.192\linewidth]{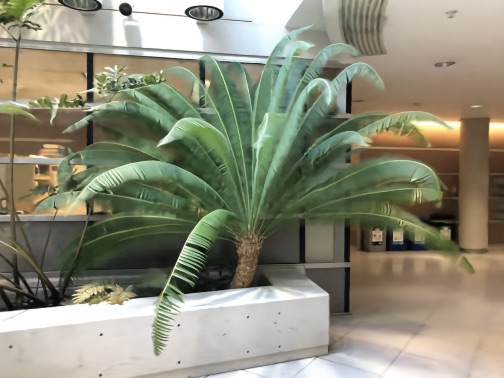}
    \\
    \includegraphics[width=0.192\linewidth]{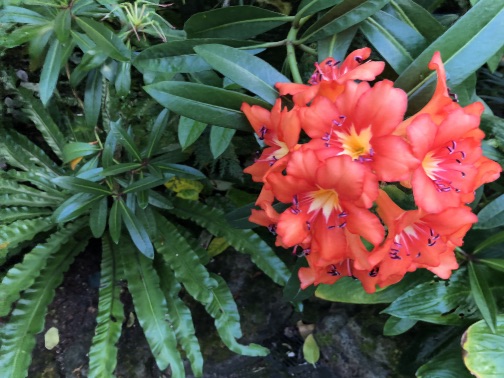}
    \includegraphics[width=0.192\linewidth]{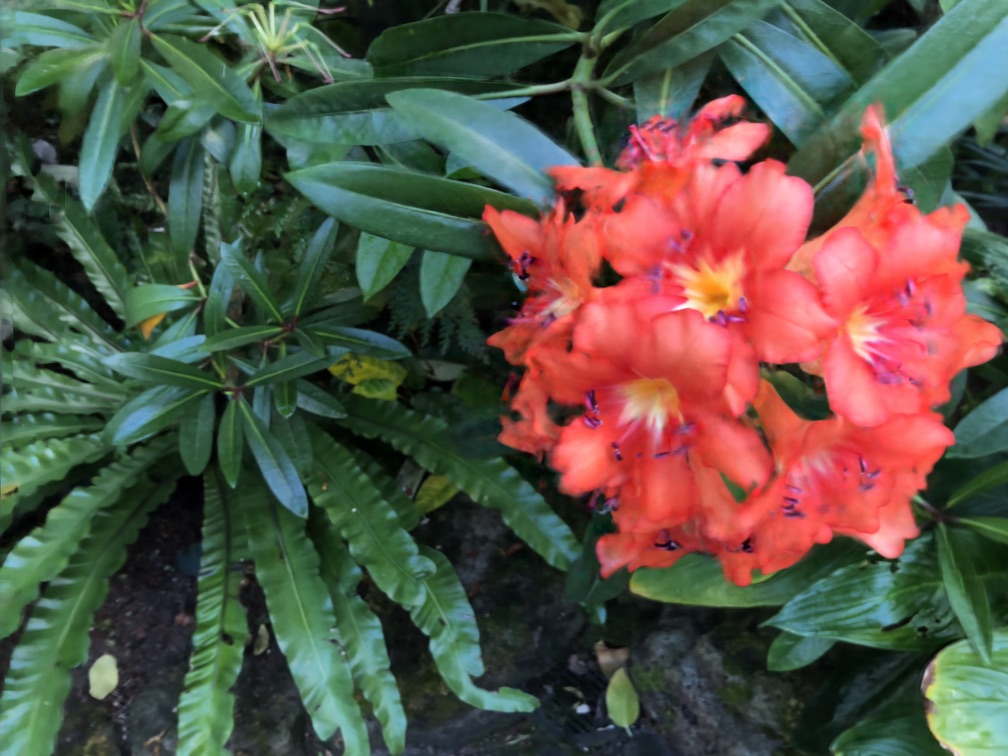}
    \includegraphics[width=0.192\linewidth]{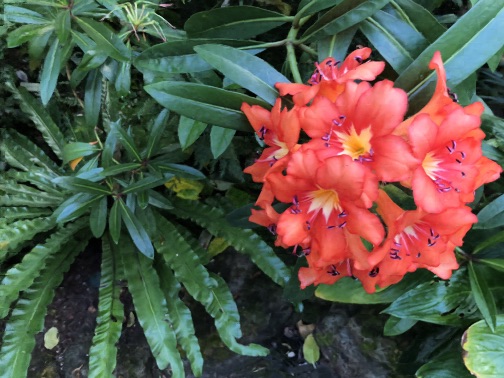}
    \includegraphics[width=0.192\linewidth]{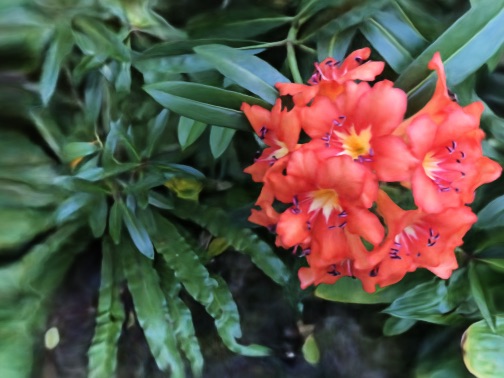}
    \includegraphics[width=0.192\linewidth]{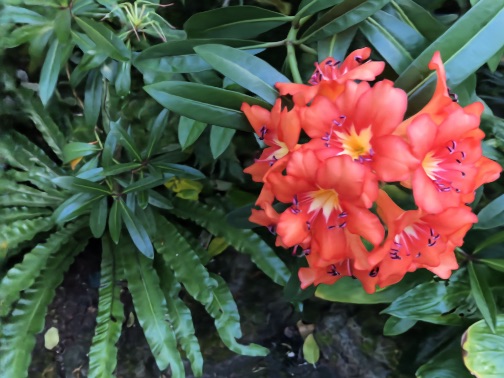}
    \\
    \includegraphics[width=0.192\linewidth]{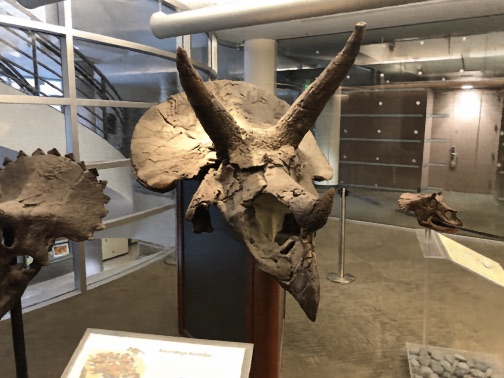}
    \includegraphics[width=0.192\linewidth]{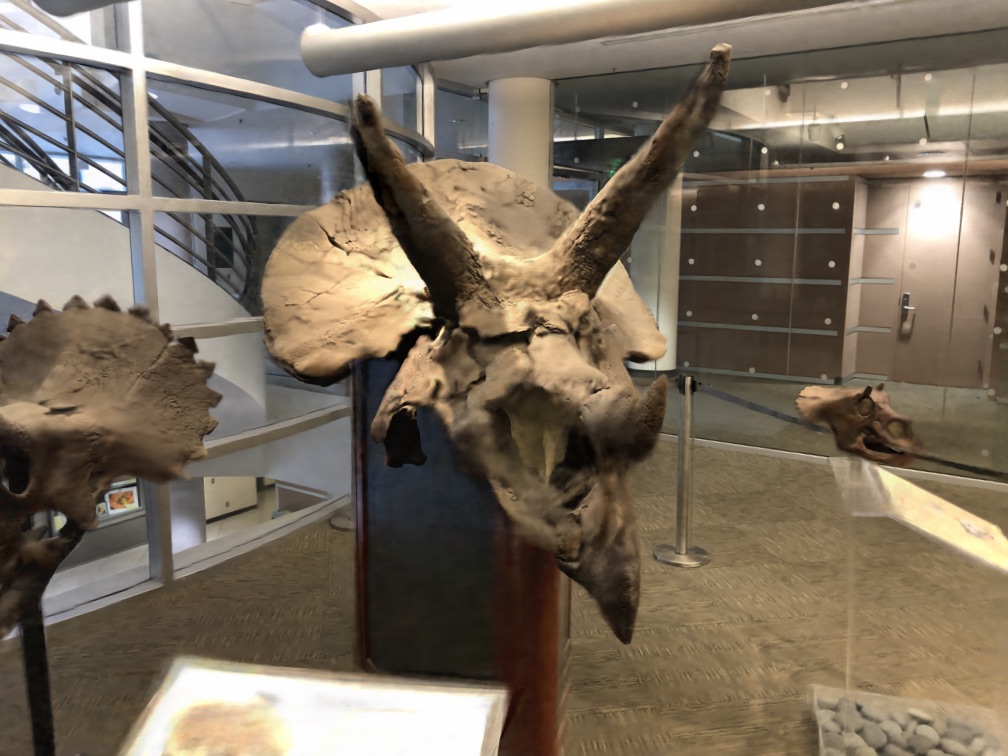}
    \includegraphics[width=0.192\linewidth]{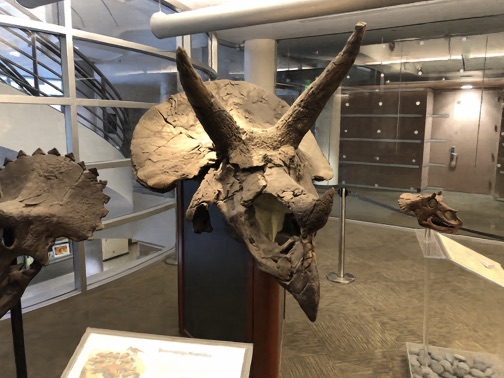}
    \includegraphics[width=0.192\linewidth]{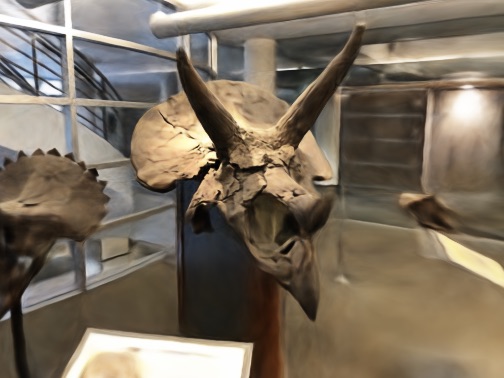}
    \includegraphics[width=0.192\linewidth]{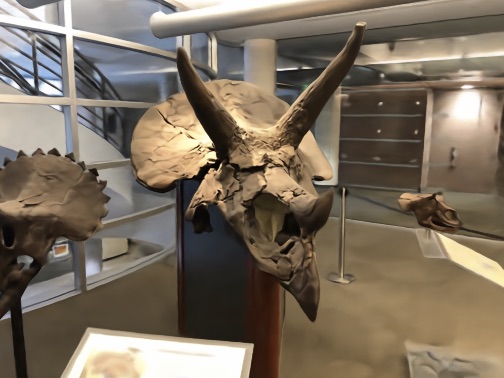}
    \\
    \includegraphics[width=0.192\linewidth]{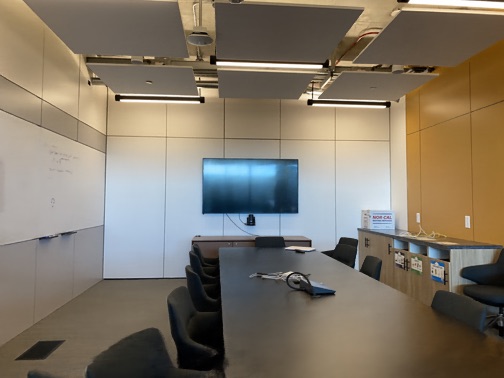}
    \includegraphics[width=0.192\linewidth]{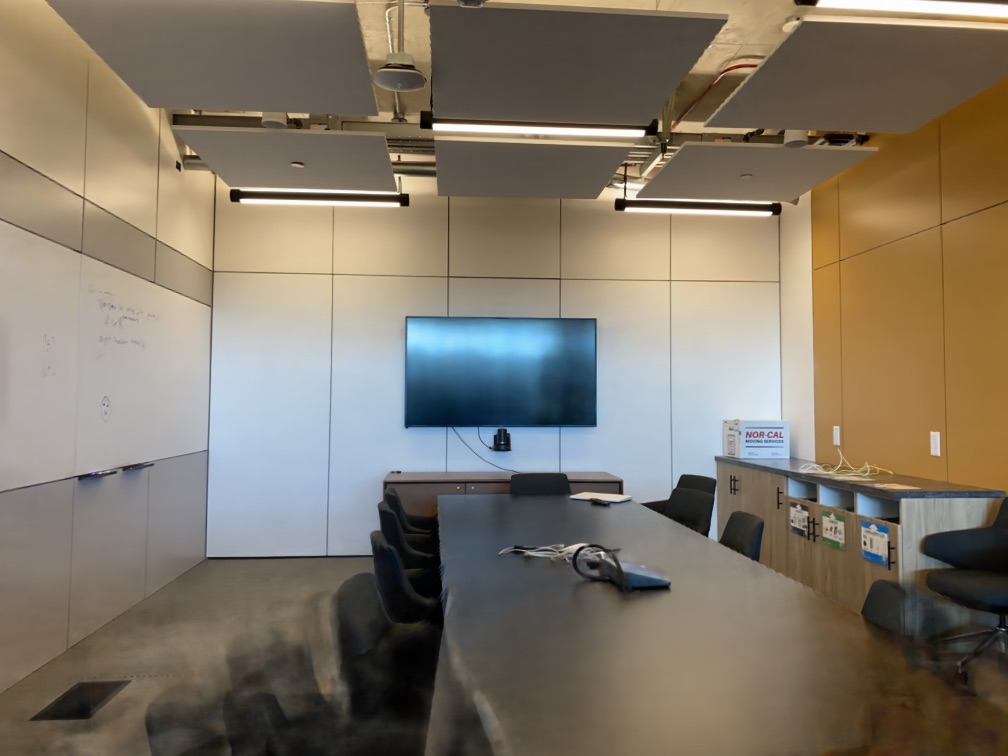}
    \includegraphics[width=0.192\linewidth]{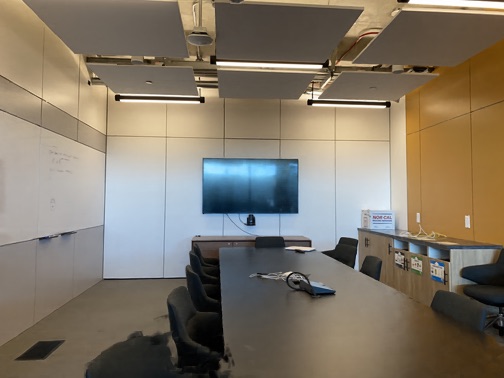}
    \includegraphics[width=0.192\linewidth]{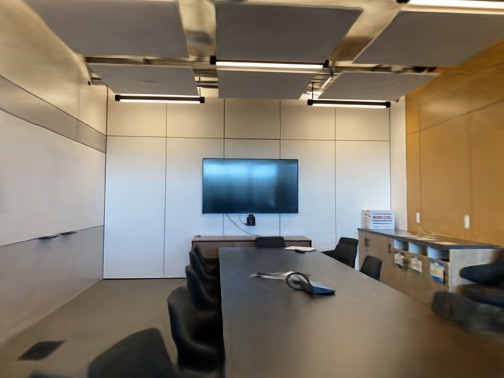}
    \includegraphics[width=0.192\linewidth]{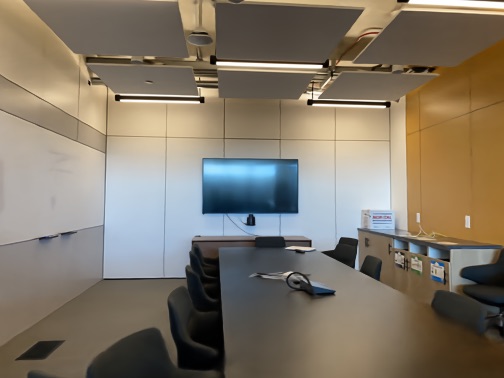}
    \\    
    \includegraphics[width=0.192\linewidth]{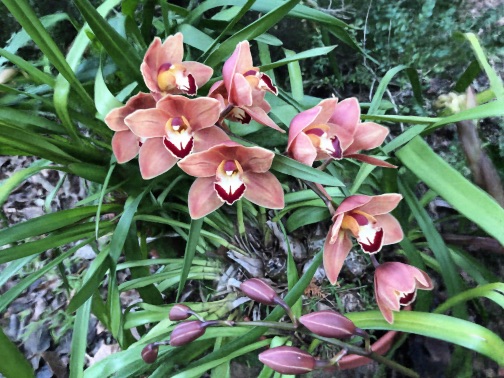}
    \includegraphics[width=0.192\linewidth]{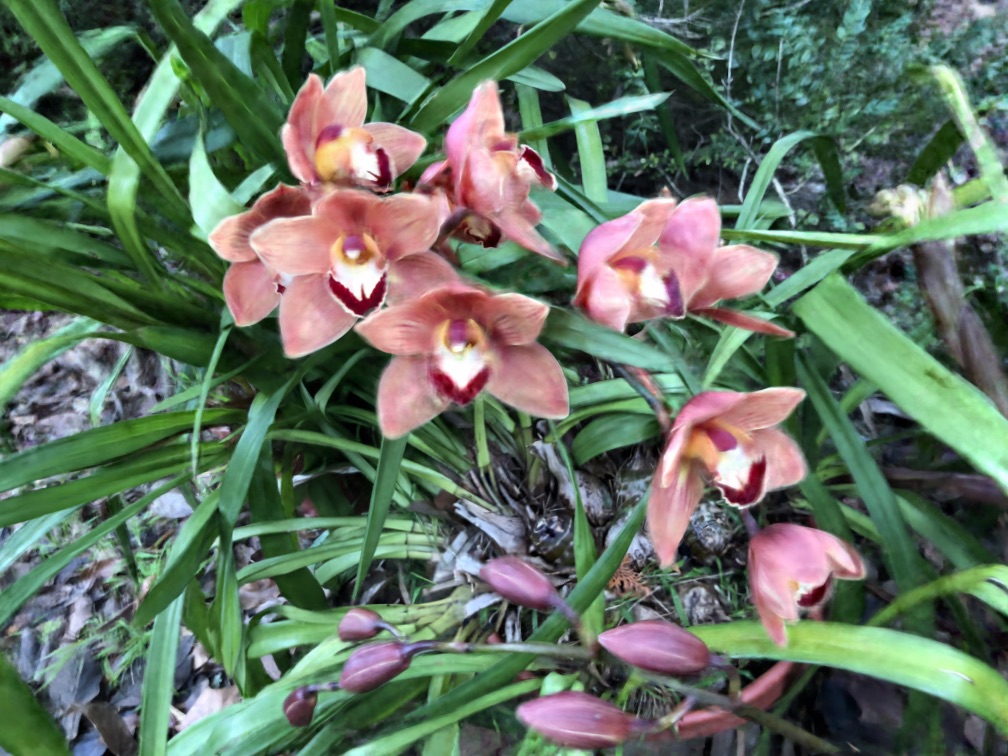}
    \includegraphics[width=0.192\linewidth]{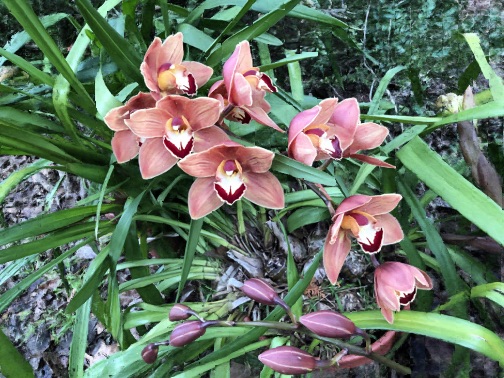}
    \includegraphics[width=0.192\linewidth]{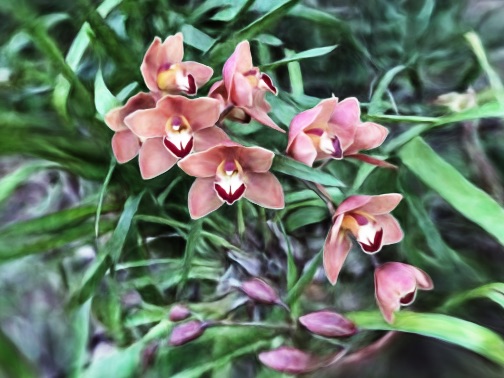}
    \includegraphics[width=0.192\linewidth]{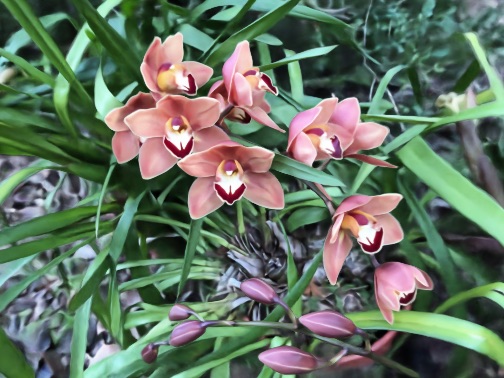}
    \\ 
    \caption{Qualitative comparisons. Compared with a 16-layer R2L (R2L-16) that produces much blur, ours has a high quality. Best viewed on a monitor. } 
    \label{fig:qualitative_compare}
    \vspace{-2mm}
\end{figure*}

%% file: tables/results_llff_nelf.tex
\begin{table}[t]
    \centering
    \caption{Quantitative comparisons with NeLF-based methods on the LLFF dataset. For methods with multiple configurations available, we report two of them which correspond to ``high-quality" and ``high-speed" respectively.}
    \label{tab:res_llff_nelf}    
    \resizebox{\linewidth}{!}{
                \begin{tabular}{@{}ccccc@{}}
        \toprule
        \multicolumn{2}{c}{Method} &
          \begin{tabular}[c]{@{}c@{}}PSNR\\  (dB)\end{tabular} &
          \begin{tabular}[c]{@{}c@{}}FPS \\ ~(RTX3090)~\end{tabular} &
          \begin{tabular}[c]{@{}c@{}}Memory Cost \\ (MB)\end{tabular} \\ \midrule
        \multicolumn{1}{c|}{\multirow{5}{*}{NeLF-based}} & R2L-16   & 24.42 & 20.12 & 4.6                  \\ \cmidrule(l){2-5} 
        \multicolumn{1}{c|}{}                            & R2L-88   & 27.79 & 4.44  & 23.0                 \\ \cmidrule(l){2-5} 
        \multicolumn{1}{c|}{}                            & RSEN-4   & 25.58 & 3.75  & \multirow{2}{*}{5.4} \\ \cmidrule(lr){2-4}
        \multicolumn{1}{c|}{}                            & RSEN-32  & 27.45 & 0.45  &                       \\ \midrule
        \multicolumn{1}{c|}{\multirow{2}{*}{Ours}}       & NeRDF-8  & 26.53 & 21.18 & 5.1                  \\ \cmidrule(l){2-5} 
        \multicolumn{1}{c|}{}                            & ~~~NeRDF-48~~~ & 27.19 & 4.39  & 28.0                 \\ \bottomrule
        \end{tabular}
    }
    \vspace{-1mm}
\end{table}

%% file: figures/fig_psnr_curve_nelf.tex
\begin{figure}[tb] 
    \centering
    \includegraphics[width=0.42\textwidth]{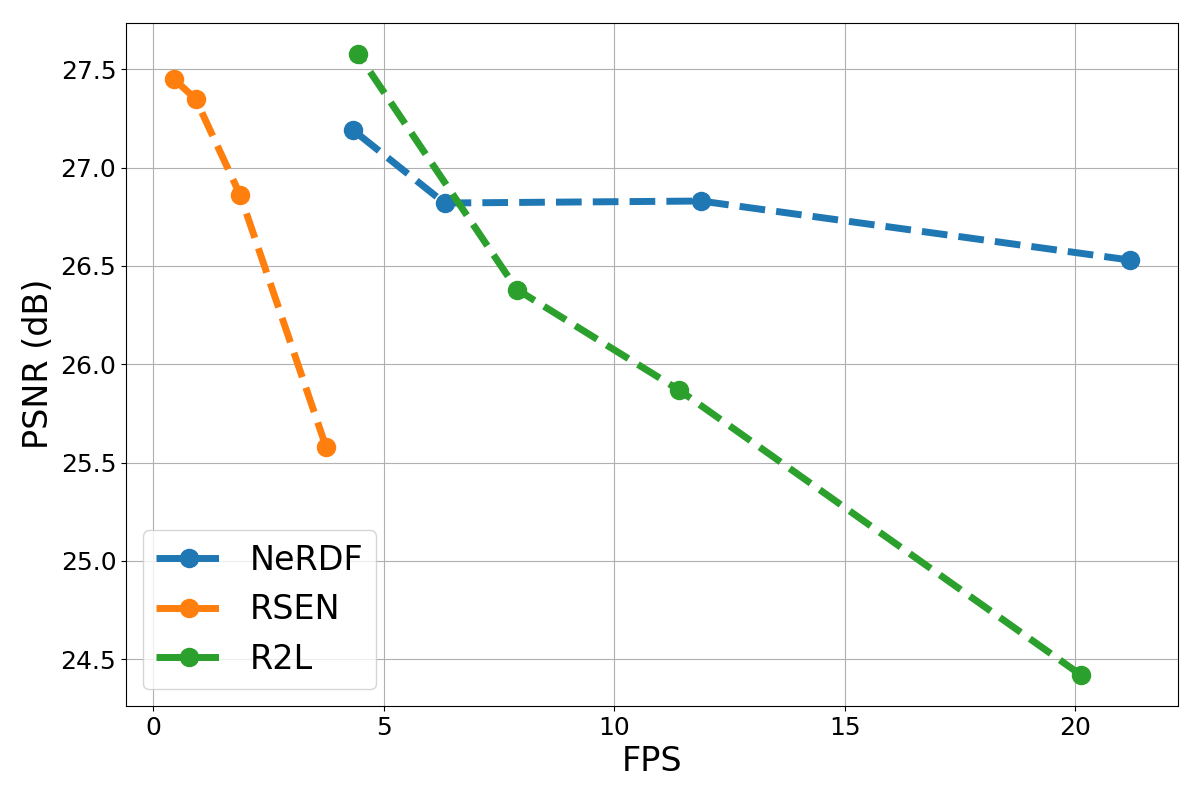}
    \caption{Trade-off curves between PSNR and FPS of NeLF-based methods and ours. Our method (the blue curve) has significant advantages in the high-FPS ($>$10) region over other NeLF-based methods.}
    \label{fig:psnr_curve_nelf}
\end{figure}

%% file: tables/results_llff_nerf.tex
\begin{table}[t]
    \centering
    \caption{Quantitative comparisons with NeRF-based methods on the LLFF dataset. For methods with multiple configurations available, we report two of them which correspond to ``high-quality" and ``high-speed" respectively. ``RTX3090*'' represents the optimized version test on an RTX3090.}
    \label{tab:res_llff_nerf}    
    \resizebox{\linewidth}{!}{
\begin{tabular}{ccccc}
\toprule
\multicolumn{2}{c}{Method} &
  \begin{tabular}[c]{@{}c@{}}PSNR\\  (dB)\end{tabular} &
  \begin{tabular}[c]{@{}c@{}}FPS \\ (RTX3090*)\end{tabular} &
  \begin{tabular}[c]{@{}c@{}}Memory Cost \\ (MB)\end{tabular} \\ \midrule
\multicolumn{1}{c|}{\multirow{13}{*}{NeRF-based}} & NeRF (Teacher) & 27.75 & 1.45      & 3.8                  \\ \cmidrule(l){2-5} 
\multicolumn{1}{c|}{}                             & TermiNeRF-2      & 21.68 & 65.49     & \multirow{2}{*}{4.1} \\ \cmidrule(l){2-4}
\multicolumn{1}{c|}{}                             & TermiNeRF-16     & 23.55 & 11.34     &                       \\ \cmidrule(l){2-5} 
\multicolumn{1}{c|}{}                             & AdaNeRF-2        & 21.82 & 101.01    & \multirow{2}{*}{4.1} \\ \cmidrule(l){2-4}
\multicolumn{1}{c|}{}                             & AdaNeRF-11       & 26.24 & 27.70     &                       \\ \cmidrule(l){2-5} 
\multicolumn{1}{c|}{}                             & DONeRF-2         & 20.89 & 65.49     & \multirow{2}{*}{4.1} \\ \cmidrule(l){2-4}
\multicolumn{1}{c|}{}                             & DONeRF-16        & 22.91 & 11.34     &                       \\ \cmidrule(l){2-5} 
\multicolumn{1}{c|}{}                             & InstantNGP-14    & 24.77 & 39.17     & 2.0                  \\ \cmidrule(l){2-5} 
\multicolumn{1}{c|}{}                             & InstantNGP-19    & 25.58 & 24.73     & 64.0                 \\ \cmidrule(l){2-5} 
\multicolumn{1}{c|}{}                             & NeX              & 27.99 & $\sim$350 & 89.0                 \\ \midrule
\multicolumn{1}{c|}{Ours}                         & NeRDF-8          & 26.53 & 369.00    & 5.1                  \\ 
\bottomrule
\end{tabular}
    }
    \vspace{-2mm}
\end{table}

%% file: figures/fig_psnr_curve_nerf.tex
\begin{figure}[tb] 
    \centering
    \includegraphics[width=0.42\textwidth]{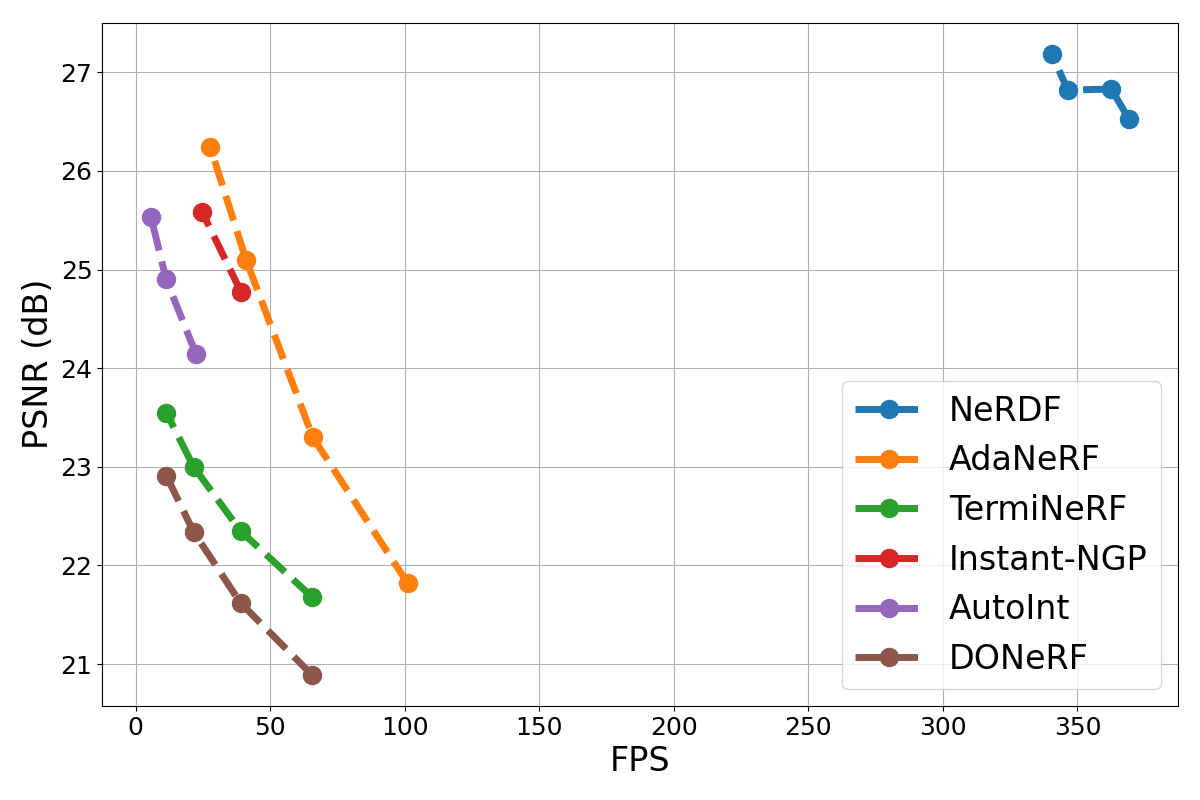}
    \caption{Trade-off curves between PSNR and FPS of NeRF-based methods and ours (both with inference optimization). Our method (the blue curve) has a significantly better trade-off than other NeRF-based methods.}
    \label{fig:psnr_curve_nerf}
\end{figure}

%% file: tables/time_analysis.tex
\begin{table}[tb]\centering
    \caption{The run time breakdown (render a 504$\times$378 image on an RTX3090 GPU)
             for each component of a NeRDF-8 with $K=12$.}
    \label{tab:time_analysis}
    \resizebox{0.95\linewidth}{!}{
        \begin{tabular}{@{}c|cccc@{}}
        \toprule
        Component & Input encoding & Network & Rendering & Total  \\
        \midrule
        Time(ms)  & 8.7 & 36.0   & 2.5       & 47.2 \\
        \bottomrule
        \end{tabular}
    }
    \vspace{-1mm}
\end{table}

%% file: tables/ablation_OVS_VDC.tex
\begin{table}[tb]\centering
    \caption{The synthesis quality (PSNR, dB) comparison between R2L and different variants of NeRDF on FERN and FLOWER.}
    \label{tab:ablation_ovs_vdc}
    \resizebox{0.95\linewidth}{!}{
    \large
    \begin{tabular}{*{6}{c}}
        \toprule
        Idx & Distribution & ~~~OVS~~~ & ~~~VDC~~~ & ~~~FERN~~~ & ~FLOWER~ \\
        \midrule
        (a) &            &            &            & 22.61 & 22.01\\
        (b) & \checkmark &            &            & 24.04 & 24.05\\
        (c) & \checkmark & \checkmark &            & 24.99 & 26.81\\
        (d) & \checkmark & \checkmark & \checkmark & \textbf{25.06} & \textbf{26.91}\\
        \bottomrule
    \end{tabular}
    }
    \vspace{-1mm}
\end{table}

%% file: figures/fig_volume_density_disparity.tex
\begin{figure}[ht!] \centering
    \makebox[0.45\linewidth]{\footnotesize w/ VDC}
    \makebox[0.45\linewidth]{\footnotesize w/o VDC}
    \\
    \includegraphics[width=0.49\linewidth]{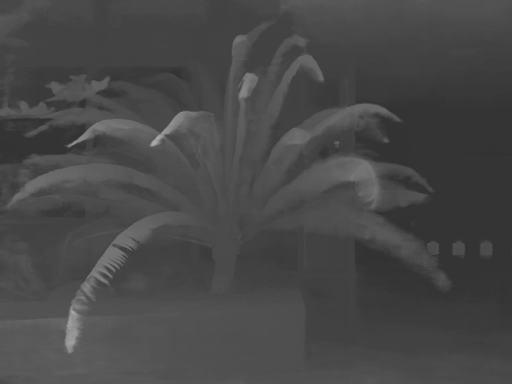}
    \includegraphics[width=0.49\linewidth]{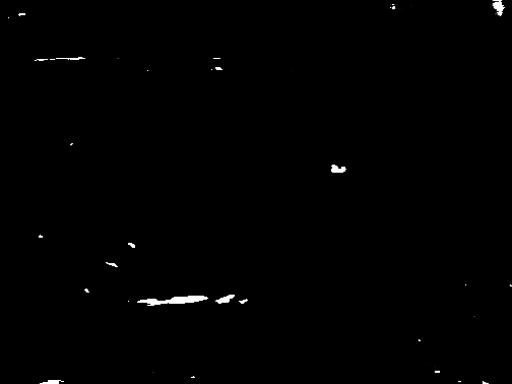}
    \\
    \caption{Visualization of synthesized disparity maps by NeRDF with and without the volume density constraint (VDC). VDC helps predict reasonable disparity.} 
    \label{fig:sigmaloss_effect_disparity}
\end{figure}

%% file: tables/ablation_K_GMM.tex
\begin{table}[tb]\centering
    \caption{The synthesis quality of using different numbers of frequencies.}
    \label{tab:ablation_K_GMM}
    \resizebox{0.9\linewidth}{!}{ \small
    \begin{tabular}{*1{c}|*5{c}}
        \toprule
        $K$ & 4 & 8 & \textbf{12} & 16 & 24 \\
        \midrule
        PSNR(dB)   & ~25.63~ & ~25.64~ & ~\textbf{25.69}~ & ~25.66~ & ~25.67~ \\
        \bottomrule
    \end{tabular}
    }
    \vspace{-1mm}
\end{table}

%% file: tables/ablation_Input_Encode.tex
\begin{table}[tb]\centering
    \caption{The synthesis quality of different variants in embedding the ray origin, direction, and on-the-path points in the input ray encoding.}
    \label{tab:ablation_input_encode}
    \resizebox{0.9\linewidth}{!}{
    \large
    \begin{tabular}{*{5}{c}}
        \toprule
        Idx & ~~~Points~~~ & ~~~Direction~~~ & ~~~Origin~~~ & ~~~PSNR(dB)~~~ \\
        \midrule
        (a) &            & \checkmark & \checkmark & 24.57 \\
        (b) & \checkmark &            &            & 25.56 \\
        (c) & \checkmark & \checkmark &            & 25.68 \\
        (d) & \checkmark &            & \checkmark & 25.57 \\
        (e) & \checkmark & \checkmark & \checkmark & \textbf{25.69} \\
        \bottomrule
    \end{tabular}
    }
    \vspace{-1mm}
\end{table}

%% file: 6_conclusions.tex
\section{Conclusion}
\label{sec:Conclusion}
We have proposed a novel 3D scene representation, Neural Radiance Distribution Field (NeRDF), for real-time efficient view synthesis. 
NeRDF models the radiance distribution along rays parameterized using a set of trigonometric functions, and synthesizes images with volumetric rendering. Our NeRDF requires only one single network forward pass per pixel for view synthesis.
NeRDF is learned by distilling both color and volume density information from a teacher NeRF.
Our method offers a better trade-off among speed, cost, and quality compared with existing efficient view synthesis methods, with a significantly high rendering speed of $\sim$350 FPS and producing visually plausible results with a small network.


%% file: 0_main.bbl
\begin{thebibliography}{10}\itemsep=-1pt

\bibitem{attal2021learning}
Benjamin Attal, Jia-Bin Huang, Michael Zollhoefer, Johannes Kopf, and Changil
  Kim.
\newblock Learning neural light fields with ray-space embedding networks.
\newblock {\em arXiv preprint arXiv:2112.01523}, 2021.

\bibitem{attal2022learning}
Benjamin Attal, Jia-Bin Huang, Michael Zollh{\"o}fer, Johannes Kopf, and
  Changil Kim.
\newblock Learning neural light fields with ray-space embedding.
\newblock In {\em CVPR}, 2022.

\bibitem{barron2021mip}
Jonathan~T Barron, Ben Mildenhall, Matthew Tancik, Peter Hedman, Ricardo
  Martin-Brualla, and Pratul~P Srinivasan.
\newblock {Mip-NeRF}: A multiscale representation for anti-aliasing neural
  radiance fields.
\newblock In {\em ICCV}, 2021.

\bibitem{barron2022mip}
Jonathan~T Barron, Ben Mildenhall, Dor Verbin, Pratul~P Srinivasan, and Peter
  Hedman.
\newblock Mip-nerf 360: Unbounded anti-aliased neural radiance fields.
\newblock In {\em CVPR}, 2022.

\bibitem{bi2020neural}
Sai Bi, Zexiang Xu, Pratul Srinivasan, Ben Mildenhall, Kalyan Sunkavalli,
  Milo{\v{s}} Ha{\v{s}}an, Yannick Hold-Geoffroy, David Kriegman, and Ravi
  Ramamoorthi.
\newblock Neural reflectance fields for appearance acquisition.
\newblock {\em arXiv preprint arXiv:2008.03824}, 2020.

\bibitem{chan2021pi}
Eric~R Chan, Marco Monteiro, Petr Kellnhofer, Jiajun Wu, and Gordon Wetzstein.
\newblock pi-gan: Periodic implicit generative adversarial networks for
  3d-aware image synthesis.
\newblock In {\em CVPR}, 2021.

\bibitem{chen2017learning}
Guobin Chen, Wongun Choi, Xiang Yu, Tony Han, and Manmohan Chandraker.
\newblock Learning efficient object detection models with knowledge
  distillation.
\newblock {\em NeurIPS}, 2017.

\bibitem{deng2022gram}
Yu Deng, Jiaolong Yang, Jianfeng Xiang, and Xin Tong.
\newblock Gram: Generative radiance manifolds for 3d-aware image generation.
\newblock In {\em CVPR}, 2022.

\bibitem{deng2021deformed}
Yu Deng, Jiaolong Yang, and Tong Xin.
\newblock Deformed implicit field: Modeling 3d shapes with learned dense
  correspondence.
\newblock In {\em CVPR}, 2021.

\bibitem{du2021_nerflow_iccv21}
Yilun Du, Yinan Zhang, Hong-Xing Yu, Joshua~B. Tenenbaum, and Jiajun Wu.
\newblock Neural radiance flow for 4d view synthesis and video processing.
\newblock In {\em ICCV}, 2021.

\bibitem{fang2021neusample}
Jiemin Fang, Lingxi Xie, Xinggang Wang, Xiaopeng Zhang, Wenyu Liu, and Qi Tian.
\newblock Neusample: Neural sample field for efficient view synthesis.
\newblock {\em arXiv preprint arXiv:2111.15552}, 2021.

\bibitem{garbin2021_fastnerf_arxiv}
Stephan~J Garbin, Marek Kowalski, Matthew Johnson, Jamie Shotton, and Julien
  Valentin.
\newblock {FastNeRF}: High-fidelity neural rendering at 200fps.
\newblock In {\em ICCV}, 2021.

\bibitem{hinton2015distilling}
Geoffrey Hinton, Oriol Vinyals, Jeff Dean, et~al.
\newblock Distilling the knowledge in a neural network.
\newblock {\em arXiv preprint arXiv:1503.02531}, 2015.

\bibitem{hong2022headnerf}
Yang Hong, Bo Peng, Haiyao Xiao, Ligang Liu, and Juyong Zhang.
\newblock Headnerf: A real-time nerf-based parametric head model.
\newblock In {\em CVPR}, 2022.

\bibitem{hu2019taichi}
Yuanming Hu, Tzu-Mao Li, Luke Anderson, Jonathan Ragan-Kelley, and Fr{\'e}do
  Durand.
\newblock Taichi: a language for high-performance computation on spatially
  sparse data structures.
\newblock {\em TOG}, 2019.

\bibitem{kurz2022adanerf}
Andreas Kurz, Thomas Neff, Zhaoyang Lv, Michael Zollh{\"o}fer, and Markus
  Steinberger.
\newblock Adanerf: Adaptive sampling for real-time rendering of neural radiance
  fields.
\newblock In {\em ECCV}, 2022.

\bibitem{li2022estimating}
Xiu Li, Xiao Li, and Yan Lu.
\newblock Estimating neural reflectance field from radiance field using tree
  structures.
\newblock {\em arXiv preprint arXiv:2210.04217}, 2022.

\bibitem{lindell2021_autoint_cvpr21}
David~B Lindell, Julien~NP Martel, and Gordon Wetzstein.
\newblock {AutoInt}: Automatic integration for fast neural volume rendering.
\newblock In {\em CVPR}, 2021.

\bibitem{liu2020_nsvf_nips20}
Lingjie Liu, Jiatao Gu, Kyaw~Zaw Lin, Tat-Seng Chua, and Christian Theobalt.
\newblock Neural sparse voxel fields.
\newblock In {\em NeurIPS}, 2020.

\bibitem{liu2021neural}
Lingjie Liu, Marc Habermann, Viktor Rudnev, Kripasindhu Sarkar, Jiatao Gu, and
  Christian Theobalt.
\newblock {Neural Actor}: Neural free-view synthesis of human actors with pose
  control.
\newblock {\em TOG}, 2021.

\bibitem{mescheder2019occupancy}
Lars Mescheder, Michael Oechsle, Michael Niemeyer, Sebastian Nowozin, and
  Andreas Geiger.
\newblock Occupancy networks: Learning 3d reconstruction in function space.
\newblock In {\em CVPR}, 2019.

\bibitem{mildenhall2019_local_tog19}
Ben Mildenhall, Pratul~P Srinivasan, Rodrigo Ortiz-Cayon, Nima~Khademi
  Kalantari, Ravi Ramamoorthi, Ren Ng, and Abhishek Kar.
\newblock Local light field fusion: Practical view synthesis with prescriptive
  sampling guidelines.
\newblock {\em TOG}, 2019.

\bibitem{mildenhall2020_nerf_eccv20}
Ben Mildenhall, Pratul~P Srinivasan, Matthew Tancik, Jonathan~T Barron, Ravi
  Ramamoorthi, and Ren Ng.
\newblock {NeRF}: Representing scenes as neural radiance fields for view
  synthesis.
\newblock In {\em ECCV}, 2020.

\bibitem{muller2022instant}
Thomas M{\"u}ller, Alex Evans, Christoph Schied, and Alexander Keller.
\newblock Instant neural graphics primitives with a multiresolution hash
  encoding.
\newblock {\em arXiv preprint arXiv:2201.05989}, 2022.

\bibitem{muller2021real}
Thomas M{\"u}ller, Fabrice Rousselle, Jan Nov{\'a}k, and Alexander Keller.
\newblock Real-time neural radiance caching for path tracing.
\newblock {\em arXiv preprint arXiv:2106.12372}, 2021.

\bibitem{neff2021_donerf_egsr21}
Thomas Neff, Pascal Stadlbauer, Mathias Parger, Andreas Kurz, Joerg~H. Mueller,
  Chakravarty R.~Alla Chaitanya, Anton~S. Kaplanyan, and Markus Steinberger.
\newblock {DONeRF: Towards Real-Time Rendering of Compact Neural Radiance
  Fields using Depth Oracle Networks}.
\newblock {\em Computer Graphics Forum}, 2021.

\bibitem{park2021_nerfies_iccv21}
Keunhong Park, Utkarsh Sinha, Jonathan~T Barron, Sofien Bouaziz, Dan~B Goldman,
  Steven~M Seitz, and Ricardo Martin-Brualla.
\newblock Nerfies: Deformable neural radiance fields.
\newblock In {\em ICCV}, 2021.

\bibitem{park2021hypernerf}
Keunhong Park, Utkarsh Sinha, Peter Hedman, Jonathan~T Barron, Sofien Bouaziz,
  Dan~B Goldman, Ricardo Martin-Brualla, and Steven~M Seitz.
\newblock {HyperNeRF}: A higher-dimensional representation for topologically
  varying neural radiance fields.
\newblock {\em TOG}, 2021.

\bibitem{passalis2018learning}
Nikolaos Passalis and Anastasios Tefas.
\newblock Learning deep representations with probabilistic knowledge transfer.
\newblock In {\em ECCV}, 2018.

\bibitem{paszke2017pytorch}
Adam Paszke, Sam Gross, Soumith Chintala, and Gregory Chanan.
\newblock {PyTorch}: Tensors and dynamic neural networks in {P}ython with
  strong {GPU} acceleration, 2017.

\bibitem{peng2021animatable}
Sida Peng, Junting Dong, Qianqian Wang, Shangzhan Zhang, Qing Shuai, Xiaowei
  Zhou, and Hujun Bao.
\newblock Animatable neural radiance fields for modeling dynamic human bodies.
\newblock In {\em ICCV}, 2021.

\bibitem{piala2021terminerf}
Martin Piala and Ronald Clark.
\newblock Terminerf: Ray termination prediction for efficient neural rendering.
\newblock In {\em 3DV}, 2021.

\bibitem{pumarola2021_dnerf_cvpr21}
Albert Pumarola, Enric Corona, Gerard Pons-Moll, and Francesc Moreno-Noguer.
\newblock {D-NeRF}: Neural radiance fields for dynamic scenes.
\newblock In {\em CVPR}, 2021.

\bibitem{Reiser2021_kiloNeRF_iccv21}
Christian Reiser, Songyou Peng, Yiyi Liao, and Andreas Geiger.
\newblock {KiloNeRF}: Speeding up neural radiance fields with thousands of tiny
  mlps.
\newblock In {\em ICCV}, 2021.

\bibitem{sitzmann2021light}
Vincent Sitzmann, Semon Rezchikov, Bill Freeman, Josh Tenenbaum, and Fredo
  Durand.
\newblock Light field networks: Neural scene representations with
  single-evaluation rendering.
\newblock {\em NeurIPS}, 2021.

\bibitem{sitzmann2019scene}
Vincent Sitzmann, Michael Zollh{\"o}fer, and Gordon Wetzstein.
\newblock Scene representation networks: Continuous 3d-structure-aware neural
  scene representations.
\newblock {\em NeurIPS}, 2019.

\bibitem{su2021nerf}
Shih-Yang Su, Frank Yu, Michael Zollh{\"o}fer, and Helge Rhodin.
\newblock A-nerf: Articulated neural radiance fields for learning human shape,
  appearance, and pose.
\newblock {\em NeurIPS}, 2021.

\bibitem{suhail2022light}
Mohammed Suhail, Carlos Esteves, Leonid Sigal, and Ameesh Makadia.
\newblock Light field neural rendering.
\newblock In {\em CVPR}, 2022.

\bibitem{te2022neural}
Gusi Te, Xiu Li, Xiao Li, Jinglu Wang, Wei Hu, and Yan Lu.
\newblock Neural capture of animatable 3d human from monocular video.
\newblock {\em arXiv preprint arXiv:2208.08728}, 2022.

\bibitem{tian2019contrastive}
Yonglong Tian, Dilip Krishnan, and Phillip Isola.
\newblock Contrastive representation distillation.
\newblock {\em arXiv preprint arXiv:1910.10699}, 2019.

\bibitem{wang2022r2l}
Huan Wang, Jian Ren, Zeng Huang, Kyle Olszewski, Menglei Chai, Yun Fu, and
  Sergey Tulyakov.
\newblock R2l: Distilling neural radiance field to neural light field for
  efficient novel view synthesis.
\newblock In {\em ECCV}, 2022.

\bibitem{wang2022fourier}
Liao Wang, Jiakai Zhang, Xinhang Liu, Fuqiang Zhao, Yanshun Zhang, Yingliang
  Zhang, Minye Wu, Jingyi Yu, and Lan Xu.
\newblock Fourier plenoctrees for dynamic radiance field rendering in
  real-time.
\newblock In {\em CVPR}, 2022.

\bibitem{wang2021neus}
Peng Wang, Lingjie Liu, Yuan Liu, Christian Theobalt, Taku Komura, and Wenping
  Wang.
\newblock {NeuS}: Learning neural implicit surfaces by volume rendering for
  multi-view reconstruction.
\newblock In {\em NeurIPS}, 2021.

\bibitem{wang2022progressively}
Peng Wang, Yuan Liu, Guying Lin, Jiatao Gu, Lingjie Liu, Taku Komura, and
  Wenping Wang.
\newblock Progressively-connected light field network for efficient view
  synthesis.
\newblock {\em arXiv preprint arXiv:2207.04465}, 2022.

\bibitem{wizadwongsa2021nex}
Suttisak Wizadwongsa, Pakkapon Phongthawee, Jiraphon Yenphraphai, and Supasorn
  Suwajanakorn.
\newblock Nex: Real-time view synthesis with neural basis expansion.
\newblock In {\em CVPR}, 2021.

\bibitem{xian2021_space_cvpr21}
Wenqi Xian, Jia-Bin Huang, Johannes Kopf, and Changil Kim.
\newblock Space-time neural irradiance fields for free-viewpoint video.
\newblock In {\em CVPR}, 2021.

\bibitem{xie2022neural}
Yiheng Xie, Towaki Takikawa, Shunsuke Saito, Or Litany, Shiqin Yan, Numair
  Khan, Federico Tombari, James Tompkin, Vincent Sitzmann, and Srinath Sridhar.
\newblock Neural fields in visual computing and beyond.
\newblock In {\em Computer Graphics Forum}, 2022.

\bibitem{yu2021plenoxels}
Alex Yu, Sara Fridovich-Keil, Matthew Tancik, Qinhong Chen, Benjamin Recht, and
  Angjoo Kanazawa.
\newblock Plenoxels: Radiance fields without neural networks.
\newblock {\em arXiv preprint arXiv:2112.05131}, 2021.

\bibitem{yu2021_plenoctrees_arxiv}
Alex Yu, Ruilong Li, Matthew Tancik, Hao Li, Ren Ng, and Angjoo Kanazawa.
\newblock Plenoctrees for real-time rendering of neural radiance fields.
\newblock In {\em ICCV}, 2021.

\bibitem{zagoruyko2016paying}
Sergey Zagoruyko and Nikos Komodakis.
\newblock Paying more attention to attention: Improving the performance of
  convolutional neural networks via attention transfer.
\newblock {\em arXiv preprint arXiv:1612.03928}, 2016.

\bibitem{zhang2021nerfactor}
Xiuming Zhang, Pratul~P Srinivasan, Boyang Deng, Paul Debevec, William~T
  Freeman, and Jonathan~T Barron.
\newblock {NeRFactor}: Neural factorization of shape and reflectance under an
  unknown illumination.
\newblock {\em TOG}, 2021.

\end{thebibliography}
